\documentclass{article}

\usepackage{arxiv}

\usepackage[utf8]{inputenc} 
\usepackage[T1]{fontenc}    
\usepackage{hyperref}       
\usepackage{url}            
\usepackage{booktabs}       
\usepackage{amsfonts}       
\usepackage{nicefrac}       
\usepackage{microtype}      
\usepackage{lipsum}		
\usepackage{graphicx}
\usepackage{natbib}
\usepackage{doi}
\usepackage[ruled]{algorithm}
\usepackage{algpseudocode}
\usepackage{amsmath}
\usepackage{amssymb} 

\usepackage{caption}
\usepackage{float} 
\usepackage{subcaption}
\usepackage{multirow}
\usepackage{multicol}
\usepackage{enumitem}
\usepackage{bbding}

\title{PolyMap: Generating High Definition Map based on Rasterized Polygons}

\author{ 
            \hspace{1mm}Shiyu Gao 
            \thanks{This work is partially derived from the author's internship at \href{https://www.qcraft.ai/en}{Qcraft Inc.}, for which we express our heartfelt gratitude.}
            \\
	Institute of Computing Technology,\\
	Chinese Academy of Science.\\
	\texttt{SibylGao1997@gmail.com} \\
	\And
        \hspace{1mm} Hao Jiang \\
	Institute of Computing Technology,\\
	Chinese Academy of Science.\\
	\texttt{jianghao@ict.ac.cn} \\
}

\renewcommand{\headeright}{Technical Report}
\renewcommand{\undertitle}{Technical Report}
\renewcommand{\shorttitle}{\textit{arXiv} Template}

\begin{document}
\maketitle

\begin{abstract}
The perception of high-definition maps is an integral component of environmental perception in autonomous driving systems. 
Existing research have often focused on online construction of high-definition maps. For instance, the Maptr\cite{liao2022maptr} series employ a detection-based method to output vectorized map instances parallelly in an end-to-end manner.
However, despite their capability for real-time construction, detection-based methods are observed to lack robust generalizability\cite{yuan2024streammapnet}, which hampers their applicability in auto-labeling systems.
Therefore, aiming to improve the generalizability, we reinterpret road elements as rasterized polygons and design a concise framework based on instance segmentation.
Initially, a segmentation-based transformer is employed to deliver instance masks in an end-to-end manner; succeeding this step, a Potrace-based\cite{selinger2003potrace} post-processing module is used to ultimately yield vectorized map elements.

Quantitative results attained on the Nuscene\cite{caesar2020nuscenes} dataset substantiate the effectiveness and generalizability of our method.
\end{abstract}


\section{Introduction}
High-definition maps (HDMap) encompass all the key static features of roads and surroundings that are necessary for autonomous driving. 
In autonomous driving systems, high-definition maps provide vital information for vehicle lane-level navigation and planning. 
Existing methods for perceiving HDMaps are mostly supervised and heavily reliant on map annotation data. The cost of annotating HDMaps, however, is significantly high. Thus, a simple yet efficient annotation model with strong generalizability can greatly enhance the efficiency of map data production.


Vectorized high-definition maps represent all instances with an ordered set of points. Open shapes such as lanes and curbs are represented by polylines, whereas closed shapes such as crosswalk are annotated as polygons.
For polylines, permutations in both forward and backward directions are considered equivalent, while for polygons, permutations starting from arbitrary vertex are deemed equivalent.

Existing DETR-based methods represent map instances as sequences of fixed number points, aiming to parallelly generate vectorized outputs, which results in:
1) \textbf{Semantic ambiguity.} Various length of each instance leads to differing densities of points.
For points, an important feature is that slight shifts within an instance do not impact their accurate representation, which current frameworks fail to capture.
2) \textbf{A propensity for overfitting.} Points that comprise an instance lack corresponding visual cues. 
As a result, the network tends to model both inter- and intra- instance relationships
with implicit attention mechanism, leading to a complex reasoning process.

Practically, we assert that in scenarios demanding generalizability, robustness, and interpretability, segmentation-based methods can demonstrate superior performance, while also enabling a more succinct and elegant framework.

It has been a longstanding suggestion by researchers to formalize lane detection as a per-pixel segmentation task\cite{chiu2005lane,neven2018towards,pan2018spatial,zheng2021resa}.
In 2021, the first online vectorized high-definition map perception network, HDMapNet\cite{li2022hdmapnet}, was proposed.
\cite{li2022hdmapnet} breaks down generation Hd map into two stages: in the first stage, semantic segmentation, instance embedding along with direction predictions are given by a LaneNet-like\cite{neven2018towards} network; in the second stage, a sophisticated post-processing module is employed to generate vectorized outputs.
Though HDMapNet\cite{li2022hdmapnet} has been found lacking in accuracy compared to detection-based methods\cite{liao2022maptr,liao2023maptrv2,liu2023vectormapnet,qiao2023end,ding2023pivotnet}, our work substantiates the potential of segmentation-based methods.
Our method achieves performance on par with detection-based methods on open datasets\cite{caesar2020nuscenes} and exhibit superior generalizability - - making our method easy to adapt into annotation systems.
Concretely, we generate a ground-truth polygon for each instance from vector annotations by rasterizing and inpainting, which are further utilized to train a transformer-based instance segmentation network. 
Under the supervision of designed rasterized ground truths, the segmentation network can output reliable segmentation results.
Considering the intrinsic elongated configuration of instances such as lanes and curbs, we introduce an inflation-based matching module to enhance the stability of instance matching, ensuring the correct allocation of positive and negative samples.
Given that the output of the network is instance-level masks, it is feasible to employ a straightforward tracing algorithm for the delineation of contours, consequently yielding vectorized results.



To summarize, the contributions of our study can be highlighted as follows:
\begin{itemize}
\item \textbf{We present a method wherein map instances are formulated as closed polygons.} 
More specifically, we construct a network for instance segmentation that end-to-end outputs rasterized map elements, followed by an vectorize post-process module based on Potrace\cite{selinger2003potrace} that outlines the contours of the instances. 
By modeling instances as polygons and generating corresponding pixel-wise labels, the performance of the instance-segmentation network is improved.
In addition, we propose an inflation-based matching module that enhances the stability of instance-matching within the transformer-based segmentation network\cite{cheng2022masked}.

\item \textbf{Superior performance on Nuscenes New Split\cite{yuan2024streammapnet}.} Experimental results on the Nuscene Old Split\cite{caesar2020nuscenes} and New Split\cite{yuan2024streammapnet} demonstrate superior generalizability of our method. 
Our method yields results on par with detection-based methods\cite{liao2022maptr,yuan2024streammapnet} on the Old Split, and surpasses them on the New Split.
\end{itemize}

\section{Related work}
\textbf{Vectorized HD map construction.}
Unlike conventional lane detection tasks, HD map perception requires output of vectorized sequences.
HDMapNet\cite{li2022hdmapnet} was the first to formalize vectorized high-precision map perception as a semantic segmentation problem, however, its accuracy has been proven to be inferior to subsequent detection-based methods\cite{liao2022maptr,liao2023maptrv2,yuan2024streammapnet,chen2024maptracker}. 
The poor performance is partly due to the rudimentary segmentation algorithm\cite{wang2018lanenet}, and partly due to the clustering-based post-processing.
To push the limit of segmentation-based methods, we introduce a transformer-based segmentation head to directly output instance masks, which significantly reduces the complexity of post-processing.
After upgrading the segmentation-based methods, we found that our method can now achieve performance close to that of the detection-based methods on the Nuscenes Old Split\cite{caesar2020nuscenes}. 
Moreover, our method also demonstrates consistent performance on the Nuscenes New Split\cite{yuan2024streammapnet}, on which the majority of Maptr\cite{liao2022maptr} series have not reported their results.

\textbf{BEV segmentation.}
Performing segmentation in the Bird's Eye View (BEV) perspective requires the initial transformation of features from frontal view to bird's eye view.
LSS\cite{philion2020lift} estimates a frustum of pesudo lidar points with semantic features from input images and acquires BEV features by pooling these points.
BEVFormer\cite{li2022bevformer} and CVT\cite{zhou2022cross} treat BEV features as learnable queries, find corresponding features on 2D images through intrinsic and extrinsic mapping and update BEV queries using complex attention mechanisms.
However, BevFormer\cite{li2022bevformer} is overly reliant on camera parameters and CVT\cite{zhou2022cross}'s consideration of global information drastically consumes computational resources. 
Consequently, GKT\cite{chen2022efficient} proposes geometry-guided kernel strategy, which extracts a small patch of features from corresponding 2D images and interacts these features with BEV queries, thereby significantly reducing computational load.
For segmentation task, most previous methods predominantly applied lightweight segmentation heads such as MLPs\cite{li2022bevformer} or 2D conv layers\cite{zhou2022cross,chen2022efficient}.
Up until now, few methods have implemented a transformer-based decoder as a segmentation head and applied it to map elements segmentation under bird's eye view.

\textbf{Tracing algorithm.}
In computer graphics, binary images can be represented as bitmaps or vector maps. Bitmaps are rasterized representations, characterizing shapes with filled grid of pixels. On the other hand, vector maps employ algebraic descriptions (such as Bezier curve) to define the outlines of an image.
The process of converting vector maps to bitmaps is called rendering, while the process of transforming bitmaps into vector maps is known as tracing.
MapVR\cite{zhang2024online} attempts to use a differentiable renderer to generate rasterized results from the vectorized sequences output by a detection-based network, and employs dice loss as auxiliary supervision. However, we found that when dice loss serves as the sole supervisory signal, it can lead to mode collapse.
In the field of font description, many traditional non-learning algorithms can efficiently perform contour tracking. However, in the realm of deep learning, achieving trainable and differentiable contour tracking remains a substantial challenge.
For the aforementioned reasons, we employ traditional tracing algorithms as a post-processing module, which can efficiently and reliably generate vectorized results.


\begin{figure}[htbp]
	\centering
	\begin{subfigure}{0.65\linewidth}
		\centering
		\includegraphics[width=1.0\linewidth]{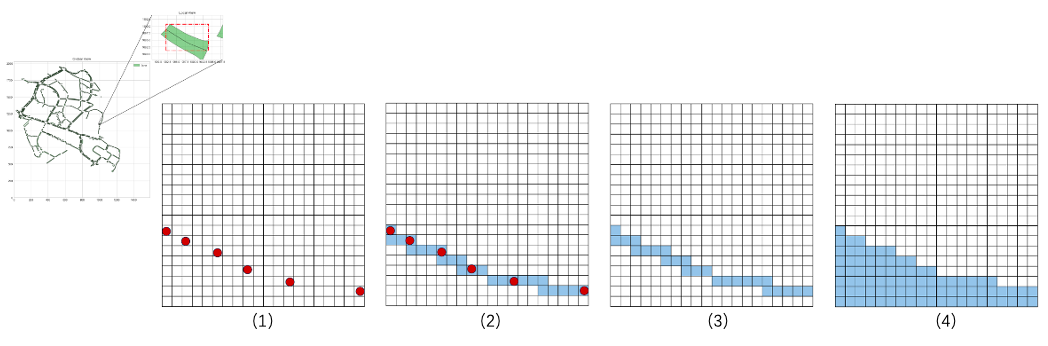}
		\caption{}
		\label{fig:ras}
	\end{subfigure}
	\centering
	\begin{subfigure}{0.31\linewidth}
		\centering
		\includegraphics[width=0.85\linewidth]{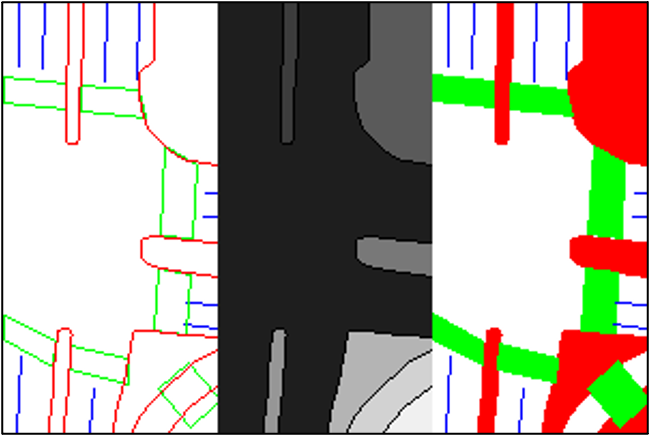}
		\caption{}
		\label{fig:ras_gts}
	\end{subfigure}
	\centering
	\caption{Rasterization process and results. a) Rasterization process. b) Rasterized polylines, connected domains, and rasterized polygons, where blue signifies lanes, green denotes ped crosses, and red represents curbs.}
	\label{fig:ras_}
\end{figure}

\section{Method}
This paper proposes a simple and straightforward approach to tackle the problem of HD map auto-labeling.
We define the problem as follows: given a set of images from surrounding views (or even raw map data in bird's eye view), our method outputs a series of vectors with high confidence scores, each of which corresponds to an instance mask on the map.
\begin{equation}
	\mathcal{M} = \mathcal{S}(\{I^{sur}_i\}^{n}_{i=1}), \quad \mathcal{V} = \mathcal{P}(\mathcal{M})
\end{equation}
wherein $\mathcal{S}(\cdot)$ represents instance segmentation network, $\mathcal{P}(\cdot)$ denotes the proposed post-processing module.
$\mathcal{M}$ and $\mathcal{V}$ are rasterized masks and vectors, respectively.

\subsection{Instance segmentation}
In accordance with most methods, we construct bev features using a query-based approach. 
Given a BEV feature (or raw BEV map data), segmenting out boundaries such as lanes and curbs is a challenging task. 
Therefore, we utilize Mask2Former\cite{cheng2022masked} as the segmentation head of our network, which can achieve higher quality instance masks compared to other methods which use lightweight segmentation heads.
Moreover, we model the road instances as polygons, and use a specific strategy to convert the annotated vector data into closed masks. 
We improve the matching method of Mask2Former\cite{cheng2022masked} to enhance the stability of the bilateral matching for certain categories, which boosts the performance of the instance segmentation network.

\subsubsection{Generate rasterized labels}
Since segmentation-based networks requires rasterized annotations, we generate instance masks based on the existing vectorized annotations in Nuscenes\cite{caesar2020nuscenes}.
Given that each category possesses distinct characteristics, we adopt different rasterization strategies for lanes, ped cross and curbs.
For lanes, we transform the vectorized annotations in the world coordinate into the BEV pixel coordinate, and fill in all pixels on the line segments between vector points, as shown in fig\ref{fig:ras}(1)(2)(3).
For ped crosses which are annotated as polygons, we initially generate rasterized line segments the same as lanes, then fill all the pixels enclosed within these line segments.
The annotations of curbs are in the form of open polylines.
However, for the segmentation network, outlining an entire region is simpler than outlining its boundary.
Moreover, the pixel area inside the curb carries consistent semantic information. 
Therefore, we consider the connected region enclosed by the curb and the boundaries of BEV image as the target for segmentation. By filling in the pixels within this connected region, we derive a semantic mask that contains information about the curb regions, as illustrated in fig\ref{fig:ras}(4).

Algorithm\ref{alg:GenPolygon} illustrates the process of generating instance masks from vectorized curb annotations.
For each curb in the scene, we first connect the vector points to generate rasterized polylines in bird's eye view.
These polylines together with image edges form connected domains.
Excluding the connected domains where the vehicle is present, each of the remaining domains is treated as a semantic label containing curb information, serving as ground truth for the curb category in the segmentation network.
Fig\ref{fig:ras_gts} illustrates the rasterized ground truth generated by the aforementioned method.

\begin{minipage}{0.45\linewidth}
\begin{algorithm}[H]
    \caption{Generate Polygon Masks for Curbs}\label{alg:GenPolygon}
    \begin{algorithmic}[1]
        \Procedure{\text{GenPolygon}}{ Set($\mathcal{V}_{curb}$) }
        \For {$v_{i}$ in Set($\mathcal{V}_{curb}$)}
            \State $\text{Draw rasterized} \ v_{i} \text{\ in Graph G}$
        \EndFor
        \State $\text{Graph} \ G_{b} \gets  \text{Binarize Graph G}$
        \State $Set(\mathcal{G}_c) \gets \text{Find connected domains in \ } G_{b}$
        \State $Set(\mathcal{G}_{curb}) = \varnothing$ 
        \For{$g_i  \in  Set(\mathcal{G}_c$)}
        \If{ $(0,0)\  \text{not in} \ g_i$ }
            \State  $Set(\mathcal{G}_{curb}) \ \cup \  g_i$
        \EndIf
        \EndFor
        \State $Set(Mask_{curb}) = \varnothing$ 
        \For{$g_i \in Set(\mathcal{G}_{curb})$}
        \State $mask_i \gets \ \text{Fill each pixels inside} \ g_i $
        \State $Set(Mask_{curb})  \ \cup \ mask_i$
        \EndFor
        \State \textbf{return} $Set(Mask_{curb})$
        \EndProcedure
    \end{algorithmic}
\end{algorithm}
\end{minipage}
\qquad 
\begin{minipage}{0.5\linewidth}
\begin{algorithm}[H]
    \small
    \caption{Vectorization Post-processing Algorithm}\label{alg:vectorize}
    \begin{algorithmic}[1]
        \Procedure{\text{Vectorization}}{ Set($\mathcal{M}$) }
        \State $Set(\mathcal{V}) \ = \ \varnothing$
        \For {${m}_{i}$ in Set($\mathcal{M}$)}
            \If{$\text{Confidence-score}({m}_{i}) \  > \ 0.5$}
                \If{$c_i \ = \ \text{"Ped Cross"}$}
                \State $\mathcal{V}_i \gets \text{Potrace}({m}_{i})$
                \EndIf
                \If{$c_i \ = \ \text{"Divider"}$}
                \State $\mathcal{V}_i \gets \text{Potrace}({m}_{i})$
                \State $\mathcal{V}_i \gets \text{RemoveLoop}(\mathcal{V}_{i})$
                \EndIf
                \If{$c_i \ = \ \text{"Curb"}$}
                \State $\mathcal{V}_i \gets \text{Potrace}({m}_{i})$
                \State $\mathcal{V}_i \gets \text{RemoveImageEdges}(\mathcal{V}_{i})$
                \EndIf
            \EndIf
            \State $Set(\mathcal{V}) \gets Set(\mathcal{V}) \ \cup \  \mathcal{V}_i$
        \EndFor
        \State \textbf{return} $Set(\mathcal{V})$
        \EndProcedure
    \end{algorithmic}
\end{algorithm}
\end{minipage}
\\
\subsubsection{Bilateral matching}

\begin{figure}[!htbp]
    \centering
    \begin{subfigure}[b]{0.23\textwidth}
      \includegraphics[width=\textwidth]{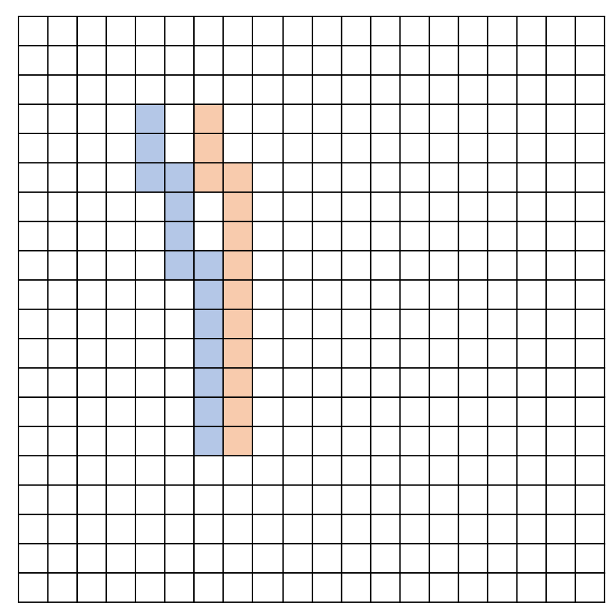}
      \caption{}
      \label{fig:match1}
    \end{subfigure}%
    \quad 
    \begin{subfigure}[b]{0.23\textwidth}
      \includegraphics[width=\textwidth]{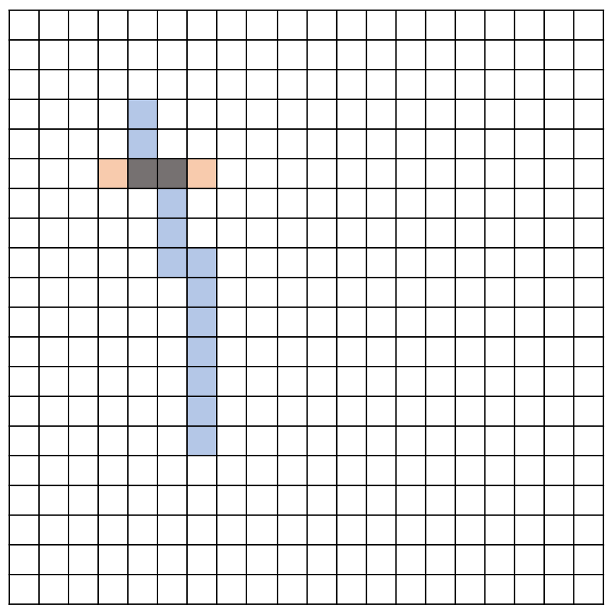}
      \caption{}
      \label{fig:match2}
    \end{subfigure}
    \quad 
    \begin{subfigure}[b]{0.23\textwidth}
      \includegraphics[width=\textwidth]{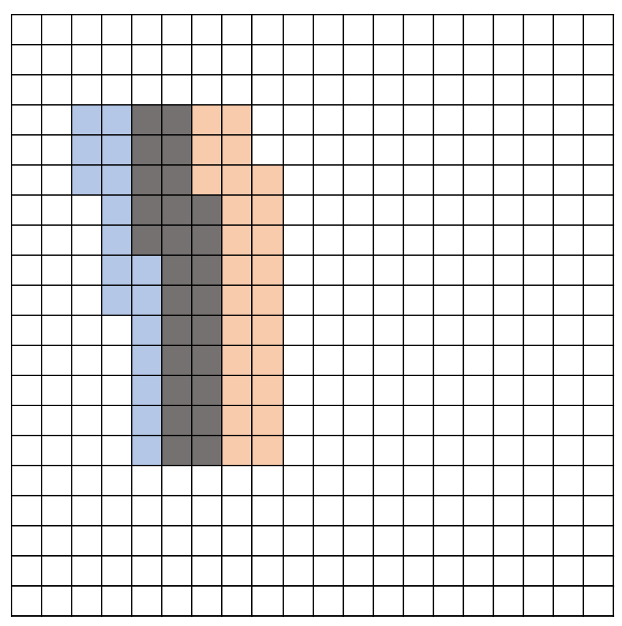}
      \caption{}
      \label{fig:match1_dil}
    \end{subfigure}
    \caption{Instance matching error. (a)mIoU=0, (b)mIoU>0, (c)Matching after dilation.}
    \label{fig:match}
\end{figure}

Mask2former calculates matching cost based on metrics like cross-entropy and intersection over union (IoU).
For narrow-shaped lanes, these metrics often fail to accurately reflect the similarity between the predictions and the ground truth.
As depicted in Fig\ref{fig:match}, blue instances represent ground truths, and orange instances represent predictions. Fig\ref{fig:match1} and Fig\ref{fig:match1_dil} each demonstrate two different situations of matching.
The predicted instance in Fig\ref{fig:match1} is supposed to be closer to the ground truth, but there is no intersection between it and the ground-truth.
In Fig\ref{fig:match1_dil}, the predicted instance is a negative sample. However, due to its overlap with the ground-truth, it is more likely to be recognised as a positive sample compared to the instance in Fig\ref{fig:match1}. 

In order to avoid the aforementioned issues, we improved the matching algorithm of Mask2former\cite{cheng2022masked} to better accommodate elongated instances such as lanes.
While many existing methods leverage signed distance function (SDF) as a metric to measure the distance between rasterized data, we found that applying a morphological dilation to both ground truths and predictions prior to bilateral matching also yields satisfactory results. 
Concretely, this study employs a convolution kernel with fixed weights in a differentiable manner to  extend the instance masks to the surrounding pixels.
After morphological dilation, the matching score between pairs of predictions and ground truths can be given by:
\begin{equation}
    \begin{cases}
    C^{ij}_{cls} \ = \  \text{Classification Cost} \ (c_i^{gt}, c_j^{p}) \\
    C^{ij}_{ce} \ = \ \text{CrossEntropy Cost} \ ( M_i^{gt}, M_j^{p} ) \\
    C^{ij}_{dice} \ = \ \text{mIoU Cost} \ (M_i^{gt}, M_j^{p})
    \end{cases}
    \label{eq:cost_3}
\end{equation}
Where classification cost indicates instance-level classification loss, with $c^{gt}$ and $c^{p}$ representing ground-truth and predicted classification results, respectively.
$m$ and $M$ denote instance masks before and after differentiable dilation, respectively.
Note that the expanded masks $M$ are used only for matching cost calculations and do not participate in loss backpropagation.
As a result, the overall matching cost is:
\begin{equation}
    C^{ij} \ = \ W^{cls} \cdot C^{ij}_{cls} + \ W^{ce}\cdot C^{ij}_{ce} + 
    \ W^{dice}\cdot C^{ij}_{dice}
    \label{eq:cost_sum}
\end{equation}

As illustrated in Fig\ref{fig:match1_dil}, after dilation, the instances in Fig\ref{fig:match1} exhibit considerable overlap, leading to a higher matching score.
The effectiveness of this method is further corroborated by empirical results on Nuscenes\cite{caesar2020nuscenes}.


\subsection{Post-processing}
We employ a rule-based post-processing module to convert binary instance masks into vectorized instances.
Unlike \cite{li2022hdmapnet}, our method outputs instance masks rather than instance embeddings in the first stage.
The high-quality instance masks enable traditional graphics methods to trace contour paths and generate approximate Bézier curves, ultimately yielding vectorized outputs.
Since most instances in this paper can be considered polygons, we chose a polygon-based tracing algorithm, Potrace\cite{selinger2003potrace}.
Given a binarized polygon, \cite{selinger2003potrace} starts from an initial point and follows predefined tracing rules along the boundary until it returns to the start point, generating a closed path.
To achieve smoother results, \cite{selinger2003potrace} approximates the path using a series of low-order Bézier curves.
In areas with greater curvature changes, more Bézier curves are assigned to fit the boundary.
Thus, the vector points generated by our method are not uniformly distributed, which we believe is advantageous for representing road instances.
After tracing, we apply certain rules to extract the desired boundaries from the polygon, as shown in Algorithm 2. 
For different categories, we use distinct post-processing logic.
For Ped Cross, the vectorization result can be obtained by directly returning the output from tracing algorithm.
For lanes, \cite{selinger2003potrace} produces a polygon that encloses the lane pixels. Thus we apply a specific algorithm to extract the centerline from closed paths.
For the curbs, we remove the points that fall on the image edges, leaving the remaining points to form the boundary of the curb.
In Section \ref{sec:exp} we will present the vectorized results for each category.

\section{Experiment}
\label{sec:exp}
\subsection{Datasets}
\textbf{Nuscenes Original Split\cite{caesar2020nuscenes}.}
Previous approaches\cite{li2022hdmapnet,liao2022maptr,liao2023maptrv2,qiao2023end,ding2023pivotnet} adhered to the Nuscenes [8] official 700/150/150 split.
However, \cite{yuan2024streammapnet} found that there is an over 84\% overlap in map locations between the training and validation sets in the original split.
This kind of overlap does not impact tasks like object detection or tracking, as different objects may appear on the same road at different times.
However, for map element perception tasks, the original split can lead to severe overfitting.
Nonetheless, most existing methods still employ the original split, ignoring the possibility that models might perform well on the validation set by memorizing the training set.

\textbf{Nuscenes New Split\cite{yuan2024streammapnet}.}
To minimize geographical overlap in map locations, \cite{yuan2024streammapnet} introduced a new data split on Nuscenes\cite{caesar2020nuscenes}. 
\cite{yuan2024streammapnet} maintained the 700/150/150 ratio for the training, validation and test sets but significantly reduced the overlap between them. 
We consider the performance on Nuscenes New Split\cite{caesar2020nuscenes} to be a more reliable indicator of a network's generalization performance.

\subsection{Metrics}
\textbf{Intersection over Union (IoU)}, the ratio of the intersection to the union between ground-truths and predictions, is a common metric in segmentation tasks.
We evaluate the segmentation results by calculating IoU, given by:
\begin{equation}
    \text{IoU}({m}_{gt}, {m}_p) \ = \ 
    \frac{\left| {m}_{gt}, \cap {m}_p \right|}{\left| {m}_{gt}, \cup {m}_p \right|}
    \label{eq:iou}
\end{equation}

\textbf{Chamfer Distance}, which is used to evaluate the distance between the ground-truth and predicted vectors, defined as:
\begin{equation}
    CD_{dir}(\mathcal{V}_1, \mathcal{V}_2) \ = \ 
    \frac{1}{\mathcal{V}_1} \sum_{x\in \mathcal{V}_1} \mathop{\min}_{y\in \mathcal{V}_2} \Vert x -y \Vert_2
    \label{eq:CD}
\end{equation}
\begin{equation}
    CD(\mathcal{V}_{gt}, \mathcal{V}_p) \ = \ 
    CD_{Dir}(\mathcal{V}_{gt}, \mathcal{V}_p) + CD_{Dir}(\mathcal{V}_p, \mathcal{V}_{gt})
    \label{eq:CD2}
\end{equation}
Where $CD_{dir}$ represents one-way chamfer distance, $CD$ denotes bidirectional. $\mathcal{V}_1$ and $\mathcal{V}_2$ are two different vectorized curves.

\subsection{Implementation}
Details omitted.

\subsection{Results and Analysis}

\begin{figure}[H]
	\centering
	\includegraphics[width=0.75\textwidth]{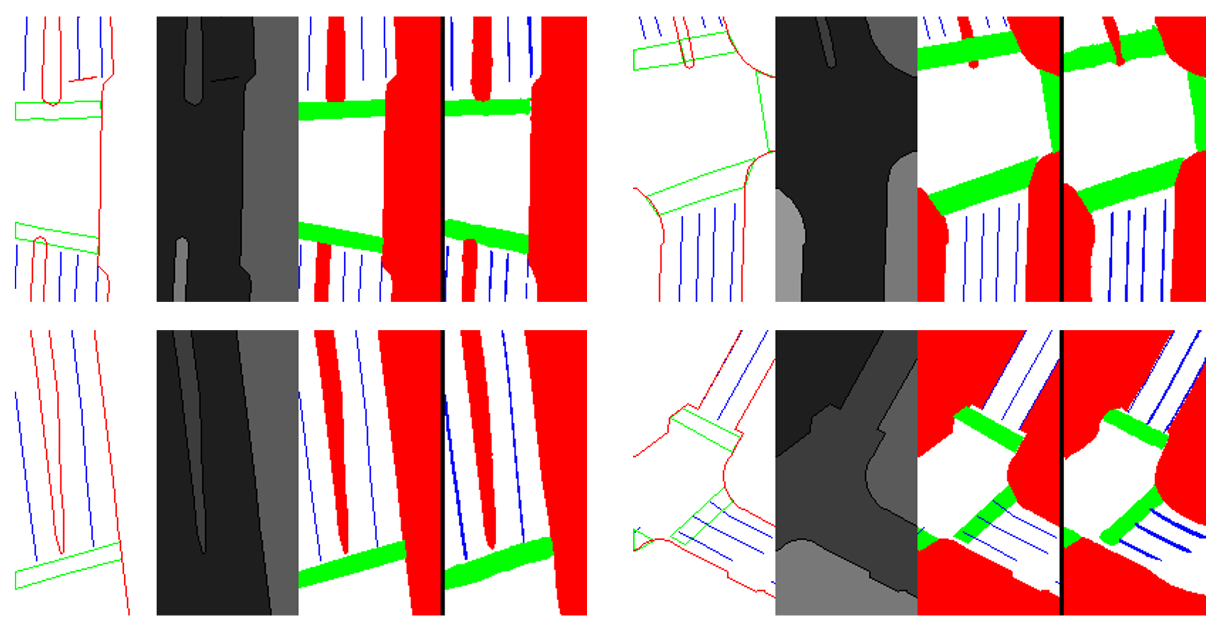}
	\caption{Segmentation results visualization. From left to right are ground-truth polyline, ground-truth connected domain, ground-truth polygon and predicted polygon mask.}
	\label{fig:seg_res}
\end{figure}

\begin{table}[!htbp]
    \caption{Quantitative results in Nuscenes Old Split.}
    \label{tab:nu_old}
    \centering
    \footnotesize
    \setlength{\tabcolsep}{4pt}
    \begin{tabular}{l|c|ccc|c|ccc|c}
    \hline
\multicolumn{1}{c|}{\multirow{2}{*}{Method}} & \multirow{2}{*}{epoch} & \multicolumn{3}{c|}{IoU $\uparrow$ } & \multirow{2}{*}{CD thr} & \multicolumn{3}{c|}{AP $\uparrow$}  & \multirow{2}{*}{mAP $\uparrow$} \\ 
\cline{3-5} \cline{7-9}
\multicolumn{1}{c|}{}  & ~ & Divider & Ped cross & Curb &                         
& Divider & Ped cross & Curb & \\ 
\hline
$\text{LSS}^{*}$\cite{philion2020lift}& 30 & 38.3 & 14.9 &  39.3  &  - & - & - & - & - \\
$\text{VPN}^{*}$\cite{pan2020cross}& 30 & 36.5 & 15.8 &  35.6  &  - & - & - & - & - \\
$\text{HDMapNet}^{*}$\cite{li2022hdmapnet}& 30 & 40.6  &  18.7   &  39.5  &  \{0.2, 0.5, 1.\} & 9.8 & 24.0 & 34.3 & 22.7 \\
$\text{Maptr}^{\ddagger}$\cite{liao2022maptr} & 24 & -  &  -   &  -  &  \{0.2, 0.5, 1.\} & 30.7 & 23.2 & 28.2 & 27.3 \\
$\text{Maptr}^{\ddagger}$\cite{liao2022maptr} & 110  & - & - & - & \{0.2, 0.5, 1.\} & \textbf{40.5} & 31.4 & \textbf{35.5} & \textbf{35.8} \\
$\text{Mapseg}^{*}$(Ours) & 24  & 43.3  &  45.8  &  42.5 &  \{0.2, 0.5, 1.\} & 25.2 & 25.2 &  22.2 & 24.2 \\
$\text{Mapseg}^{*}$(Ours) & 150  & \textbf{53.9}  &  \textbf{55.9}   &  \textbf{48.8} &  \{0.2, 0.5, 1.\} & 32.3 & \textbf{32.5} &  25.3 & 30.0 \\
\hline
\hline
$\text{HDMapNet}^{*}$\cite{li2022hdmapnet}& 30 & 40.6  &  18.7   &  39.5  &  \{0.5, 1.0, 1.5\} & 14.4 & 21.7 & 33.0 & 23.0 \\
$\text{VectorMapNet}^{\ddagger}$\cite{liu2023vectormapnet}& 110 & -  &  -   &  -  &  \{0.5, 1.0, 1.5\} & 36.1 & 47.3 &  39.3 & 40.9 \\
$\text{Maptr}^{\ddagger}$\cite{liao2022maptr} & 24 & -  &  -   &  -  &  \{0.5, 1.0, 1.5\} & 46.3 & 51.5 & 53.1 & 50.3  \\
$\text{Maptr}^{\ddagger}$\cite{liao2022maptr} & 110 & -  &  -   &  -  &  \{0.5, 1.0, 1.5\} & \textbf{56.2} & \textbf{59.8} & \textbf{60.1} & \textbf{58.7} \\
$\text{Mapseg}^{*}$(Ours) & 24  & 43.3  &  45.8  &  42.5 &  \{0.5, 1.0, 1.5\} & 34.0 & 40.5 &  44.5 & 39.7 \\
$\text{Mapseg}^{*}$(Ours) & 150  & \textbf{53.9}& \textbf{55.9} & \textbf{48.8} & \{0.5, 1.0, 1.5\} & 51.2 & 51.8 &  51.3 & 51.4 \\
\hline
\end{tabular}
\vspace*{3ex}
\end{table}

\begin{table}[!htbp]
    \caption{Quantitative results in Nuscenes New Split.}
    \label{tab:nu_new}
    \centering
    \footnotesize
    \setlength{\tabcolsep}{4pt}
    \renewcommand{\arraystretch}{0.9}
    \begin{tabular}{l|c|ccc|c|ccc|c}
    \hline
\multicolumn{1}{c|}{\multirow{2}{*}{Method}} &\multirow{2}{*}{epoch}& \multicolumn{3}{c|}{IoU $\uparrow$} & \multirow{2}{*}{CD thr} & \multicolumn{3}{c|}{AP $\uparrow$}  & \multirow{2}{*}{mAP $\uparrow$} \\ 
\cline{3-5} \cline{7-9}
\multicolumn{1}{c|}{}  & ~ & Divider & Ped cross & Curb &                         
& Divider & Ped cross & Curb & \\ 
\hline
VectorMapNet\cite{liu2023vectormapnet} & 24 & -  &  -   &  -  &  \{0.5, 1.0, 1.5\} & 40.5 & 31.4 & 35.5 & 35.8 \\
Maptr\cite{liao2022maptr} & 24 & -  &  -   &  -  &  \{0.5, 1.0, 1.5\} & 6.4 & 20.7 & 35.5 & 20.9  \\
$\text{StreamMapNet}^{\dagger}$\cite{liao2022maptr} & 24  & -  &  -   &  -  &  \{0.5, 1.0, 1.5\} &29.6 &30.1 &41.9 &33.9 \\
Mapseg(Ours) & 24  & 34.0  &  40.4  &  44.5  & \{0.5, 1.0, 1.5\} & 34.0 & 40.5 &  44.5 & 39.7 \\
Mapseg(Ours) & 150  & 52.5  &  54.3   &  50.6  &  \{0.5, 1.0, 1.5\}  & \textbf{51.2} & \textbf{51.8} &  \textbf{51.4} & \textbf{51.5}  \\
\hline
\hline
Mapseg(Ours) & 24  & 34.0  &  40.4  &  44.5  &  \{0.2, 0.5, 1.\} & 19.4 & 21.5 &  23.9 & 21.6 \\
Mapseg(Ours) & 150  & 52.5  &  54.3   &  50.6 &  \{0.2, 0.5, 1.\} & 32.1 & 29.8 &  29.5 & 30.4 \\
\hline
\end{tabular}
\vspace*{3ex}
\end{table}

\begin{table}[!htbp]
    \caption{Ablation study on Nuscenes validation set.}
    \label{tab:map_ablation}
    \centering
    \footnotesize
    \setlength{\tabcolsep}{3pt}
    \renewcommand{\arraystretch}{1.2}
    \begin{tabular}{l|c|c|c|cccc|cccc}
    \hline
\multirow{2}{*}{Method} & \multirow{2}{*}{epoch} & \multirow{2}{*}{Polygon} & \multirow{2}{*}{Dilation}&\multicolumn{4}{c|}{CD\{0.2, 0.5, 1\}} & \multicolumn{4}{c}{CD\{0.5, 1.0, 1.5\}} \\
\cline{5-12}
 & &  &  & Divider   & Ped cross  & Curb  & mAP  & Divider   & Ped cross   & Curb   & mAP  \\
\hline
Mapseg-polyline & 150 & \XSolid  & \XSolid  & 27.1  & 31.9 & 2.9 &  20.6 & 44.5 &  \textbf{55.9}& 8.0 &  36.1\\
Mapseg-wo-dil & 150 & \checkmark  & \XSolid  & 29.6  & 30.4 & \textbf{26.3} &  28.7 & 47.0 &  52.0& 47.5 &  48.8\\
Mapseg & 150 & \checkmark  & \checkmark  & \textbf{32.3} & \textbf{32.5} &  25.3 & \textbf{30.0} & \textbf{51.2} & 51.8 &  \textbf{51.3} & \textbf{51.4}\\
\hline
\end{tabular}
\end{table}

\bibliographystyle{plain}
\bibliography{references}  







\end{document}